\documentclass[preprint,12pt,authoryear]{elsarticle}

\usepackage{amssymb}
\journal{Internet of Things}

\begin{document}

\begin{frontmatter}



\title{Federated Learning on Edge Sensing Devices: A Review}


\author[label]{Berrenur Saylam}, \ead{berrenur.saylam@boun.edu.tr}
\author[label]{Özlem Durmaz İncel} \ead{ozlem.durmaz@boun.edu.tr}

%\affiliation[label]{organization={Computer Engineering Department, Boğaziçi University}, addressline={Bebek},  city={İstanbul}, postcode={34342},    country={Türkiye}}

\begin{abstract}
The ability to monitor ambient characteristics, interact with them, and derive information about the surroundings has been made possible by the rapid proliferation of edge sensing devices like IoT, mobile, and wearable devices and their measuring capabilities with integrated sensors. Even though these devices are small and have less capacity for data storage and processing, they produce vast amounts of data. Some example application areas where sensor data is collected and processed include healthcare, environmental (including air quality and pollution levels), automotive, industrial, aerospace, and agricultural applications. These enormous volumes of sensing data collected from the edge devices are analyzed using a variety of Machine Learning (ML) and Deep Learning (DL) approaches. However, analyzing them on the cloud or a server presents challenges related to privacy, hardware, and connectivity limitations. Federated Learning (FL) is emerging as a solution to these problems while preserving privacy by jointly training a model without sharing raw data. In this paper, we review the FL strategies from the perspective of edge sensing devices to get over the limitations of conventional machine learning techniques. We focus on the key FL principles, software frameworks, and testbeds. We also explore the current sensor technologies, properties of the sensing devices and sensing applications where FL is utilized. We conclude with a discussion on open issues and future research directions on FL for further studies.

\end{abstract}

\begin{graphicalabstract}
\centering
\includegraphics[width=\columnwidth]{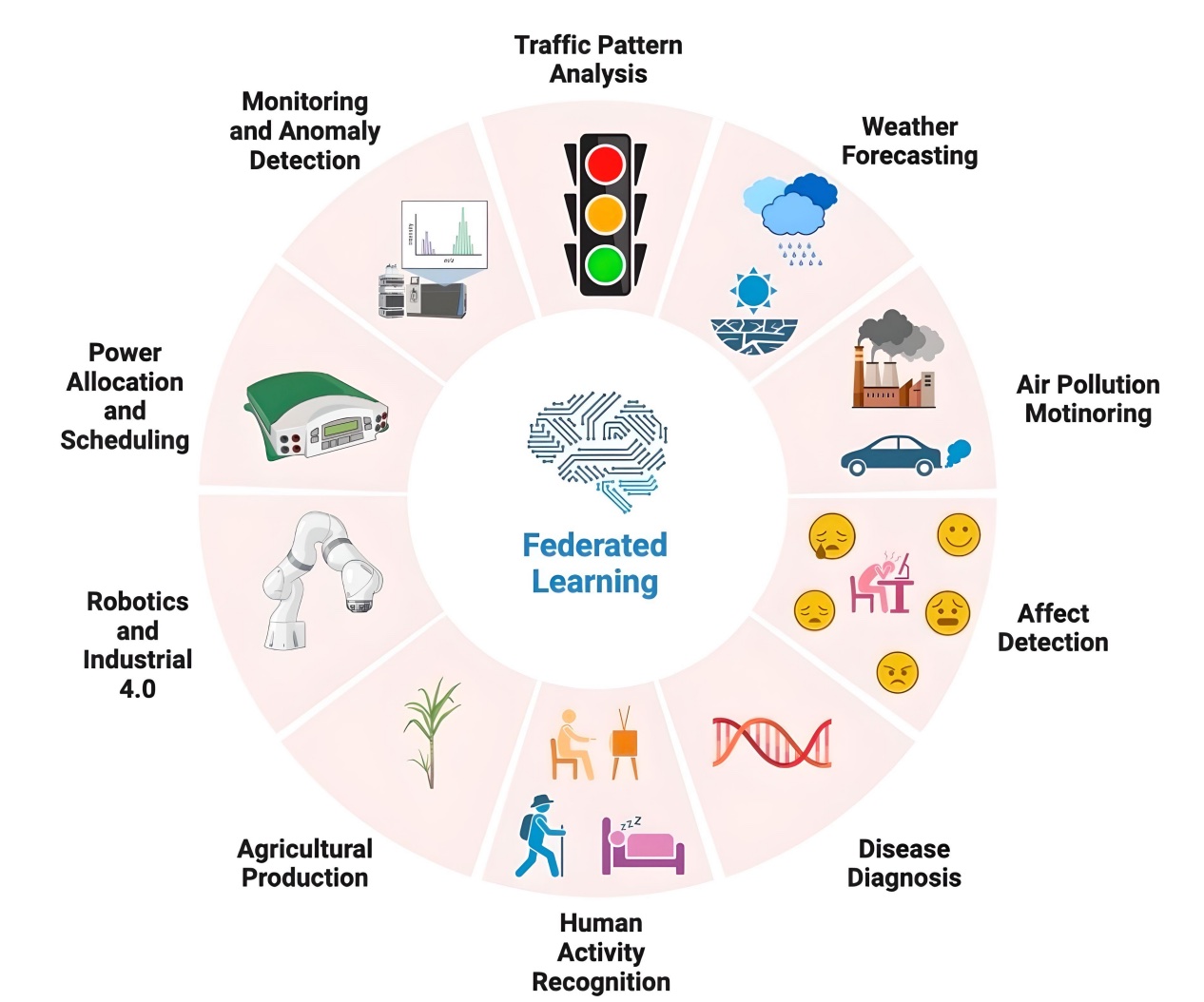}
\end{graphicalabstract}

\begin{highlights}
\item \textit{Cutting-Edge Perspective:} Our manuscript, titled ``Federated Learning on Edge Sensing Devices: A Review," provides an up-to-date and comprehensive exploration of federated learning (FL) within the context of edge sensing devices. It offers a fresh perspective on this rapidly evolving field, showcasing its relevance to the contemporary landscape of data analysis and privacy preservation.

\item \textit{Addressing Critical Challenges:} We address the pressing challenges associated with traditional centralized data analysis methods, such as privacy concerns, hardware constraints, and connectivity limitations. By highlighting the limitations of conventional approaches, our research underscores the urgency of adopting FL techniques to unlock the full potential of edge sensing devices.

\item \textit{Wide-Ranging Application Domains:} The manuscript sheds light on the broad spectrum of application areas where edge sensing devices play a pivotal role, including healthcare, environmental monitoring, automotive technology, industrial processes, aerospace, and agriculture. Our work emphasizes how FL can be a game-changer in extracting meaningful insights from sensor data in these critical domains.

\item \textit{Comprehensive Review:} We offer a comprehensive review of FL strategies tailored specifically for edge-sensing devices. This review encompasses the fundamental principles of FL, software frameworks for implementation, and practical testbeds. It serves as a valuable resource for researchers and practitioners seeking to navigate the complex landscape of FL in edge computing.

\item \textit{Current Sensor Technologies:} Our manuscript dives into the current state of sensor technologies, providing an in-depth exploration of the properties and capabilities of these devices. This insight is crucial for understanding the foundations upon which FL operates and how it can be effectively integrated into sensor-driven applications.

\item \textit{Future Research Directions:} By identifying open issues and charting future research directions, our work contributes to the ongoing discourse in the field. We invite further investigation into the potential of FL in edge sensing devices, offering a roadmap for researchers to explore uncharted territories and address emerging challenges.

\item \textit{Timely and Substantial Contribution:} Given the increasing importance of FL in the era of edge computing, our manuscript offers a timely and substantial contribution to the journal's mission of advancing knowledge in the realms of diagnostics, human health, well-being, and activity recognition with wearable sensors.

\item \textit{Unpublished and Unbiased:} We assure the journal that neither this manuscript nor any of its contents are under consideration or published in another journal, demonstrating our commitment to presenting original and unbiased research.
\end{highlights}


\end{frontmatter}


\section{Introduction}
Sensor-integrated devices, such as IoT devices, wearables, smartphones, drones, and robots, make it possible to collect vast amounts of sensing data to learn more about or extract knowledge from the environment and the surrounding users. Such edge devices can have varied resources regarding memory, battery capacity, and computing power. Even though some devices may appear more potent regarding computing capabilities, they are still considered edge devices because extra resources might not be provided \cite{a1}. 

Many use cases and application areas utilize the sensing data from these edge devices. For example, GPS data from a phone can be used to determine how congested the local traffic is, or motion-sensor data from a wearable can be used to determine a person's daily activity patterns \cite{a2}. Mainly, machine learning techniques are commonly applied to sensing data. The total amount of collected data has increased as these devices are used by masses and connected via wireless interfaces, making it easy to transfer data for further processing. In traditional learning methods, data is usually gathered in one location on a server-type machine, and then learning algorithms are trained and applied to the aggregated data.

A typical machine learning pipeline involves gathering data from various sensors to extract a significant result following an objective. However, several challenges arise for sensor data collection and processing with a variety of devices. Since different devices have different hardware, even sensor data acquisition may differ. There may also be differences in the volume of data gathered, the data sampling frequencies, the battery capacity of the devices, and the cost of communication used to transfer the data to the central data processing unit. 

It is also challenging to get meaningful results without sufficient, high-quality data, which is often the case with sensor data. For example, each device may not capture the same data distribution due to its placement or sensing trajectory. For instance, the human activity recognition field corresponds to classifying performed activity types. Since various people may have different characteristics, these types may vary amongst individuals. As a result, learning algorithms' outcomes, mainly supervised ones, may be skewed because the training data does not contain a diverse set of classes. This requirement forces building a model with numerous classes, preferably with a significant amount of data for each class.

Besides, analyzing the gathered data at one location brings another challenge: typically, this raises a privacy concern. The GDPR (General Data Protection Regulation \cite{a3}) is a recently-introduced example law to secure users' data. Additionally, the collection of data at one location increases the expense of communication. Furthermore, applying algorithms as a global model may not be effective from a security perspective since the whole system can be corrupted with one successful cyber attack, especially for medical domains. Therefore, gathering a large volume of data in one location is not practical.  

An alternative to centralized sensor data processing is to run the machine learning models directly on edge devices. However, edge devices have hardware limitations that reduce their ability to execute complex learning algorithms, particularly deep architectures. Deep learning algorithms require higher processing power and energy in addition to memory. Consequently, we require new mechanisms to adapt the centralized and on-device models to modest computation restrictions~\cite{a4}. Considering that resource efficiency and privacy are the two critical factors in many sensing applications for edge devices, combining data from many sources while maintaining data privacy is necessary to produce resource-efficient learning systems with such devices.

A promising paradigm for cooperative and privacy-preserving machine learning in dispersed contexts is federated learning (FL)~\cite{a9}. FL has attracted considerable attention in several disciplines since it allows training on local data while keeping it private and decentralized. Google presented FL in 2016 as a solution to these problems while protecting user privacy. The goal is to jointly train a model on each device without sharing raw data and produce meaningful results by only sharing model parameters with a central orchestrating unit \cite{a5}.

This paper aims to explore Federated Learning, a novel approach to learning to provide a state-of-the-art assessment from the viewpoint of sensor-based edge devices and provide an overview of FL applications on sensing devices. In literature, some survey studies \cite{a22, a102} focus on federated learning with IoT devices. In this article, we examine FL from a broader perspective including sensing edge devices, such as IoT sensors, wearables, and other mobile devices. Most current surveys~\cite{a22, a23, a28, a30} on FL offer overviews of one specific application case without providing the overall state in the literature.

Sensors are measuring devices used in various applications, including medical, environmental, automotive, industrial, aerospace, and agriculture. They detect changes in their environment, enabling precise monitoring, control, and decision-making processes. Medical sensors like ECGs and pulse oximeters help diagnose heart conditions, while environmental sensors assess water quality, pH, and dissolved oxygen. Automotive applications use light sensors, rain sensors, flow sensors, vibration sensors, and gyroscopes for stability and navigation accuracy. The diverse range of sensors used in various domains contributes to producing multimodal sensor data.

In this paper, we revisit this hot topic from the standpoint of boosting sensing devices, considering the practice of many sensing domains. Our contribution is divided into two parts: a comprehensive review of the state of the art on FL, including major sensing application domains, and state-of-the-art frameworks along with testbeds for the application of FL, which are not generally included in other surveys. Furthermore, we provide extensive discussions that are prominent for further studies.

\begin{figure}[!t]
\centerline{\includegraphics[width=0.7\columnwidth]{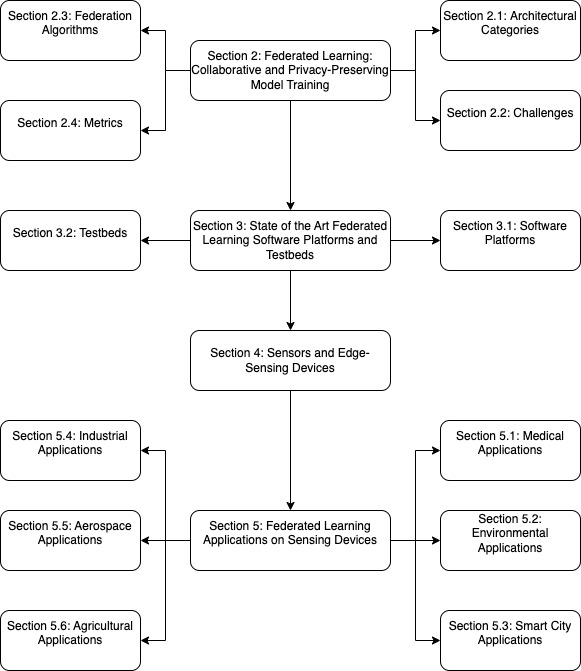}}
\caption{Schematic of the review  \label{fig:akis}}

\end{figure}

%The study is structured as follows
The structure of this paper is shown in Figure \ref{fig:akis}. In Section \ref{sec:fl}, we describe the FL concept and working procedure, along with its primary challenges, methods, and measurements. We introduce the most cutting-edge platforms for using FL algorithms and a collection of testbeds to provide applicability and encourage future studies in Section \ref{sec:hardSoft}. Section \ref{sec:sensorDevices} describes the different kinds of sensors and devices and their properties. In Section \ref{sec:flApp}, we put together the application areas and current usage of FL techniques in real-life. We discuss the state-of-the-art approaches for sensing devices, suggest further directions in Section \ref{sec:discussion}, and conclude with Section \ref{sec:conclusion}.

\section{Federated Learning: Collaborative and Privacy-Preserving Model Training}
\label{sec:fl}

This section delves into the field of FL, a method created to produce accurate models while protecting the privacy of people and institutions. As mentioned, it proposes a collaborative learning paradigm in which numerous devices take part in training a shared model under the direction of a central server. This section investigates the nuances of this collaboration procedure and its various architectural strategies.

\begin{figure}[!t]
\centerline{\includegraphics[width=0.8\columnwidth]{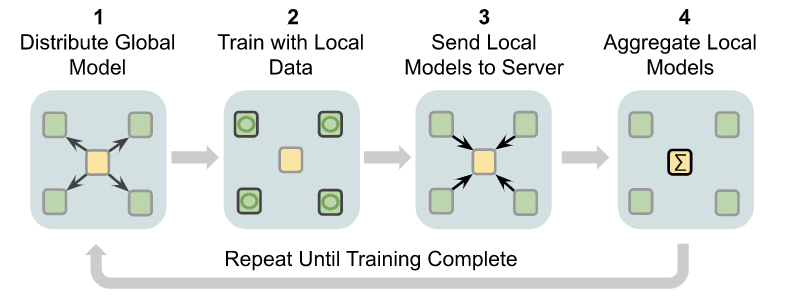}}
\caption{Typical federated learning workflow  \cite{a7}\label{fig:FL}}
\label{fig:TypicalFL}
\end{figure}

\textbf{Collaborative Model Training:} FL is a notion that respects privacy. Firstly the server distributes a model with randomly initialized parameters to the participating devices. Each device uses its local data to train the shared model, ensuring no raw data leaves the device. Instead, the server receives only the modified model parameters. The model averaging method is often used to reach a stopping point in this iterative procedure \cite{a7, a8} (a visual scheme of the process can be seen in Figure \ref{fig:FL}). By aggregating the updated parameters from all devices, a global model is obtained, embodying the collective knowledge of the participants while preserving data privacy.

\subsection{Architectural Categories}
FL encompasses three main architectural categories: horizontal, vertical, and transfer federated learning (FTL) \cite{a9}. These categories are determined based on the data distribution and the collaboration's nature. In the horizontal architecture, data features are similar across devices, with variations primarily in data volume. On the other hand, the vertical architecture focuses on data with similar data correspondence (IDs) but differing feature spaces. Horizontal architecture, often called sample-based FL, is commonly employed in medical cases where different local devices share similar data structures.
In contrast, vertical architecture, known as feature-based FL, is well-suited for aggregating user data from different servers. This architecture enables communication and collaboration between devices, requiring aggregating diverse features \cite{a10}. An example of horizontal architecture is the Google keyboard, where users from different regions update model parameters locally, which are then aggregated by the server. In contrast, the vertical architecture finds its relevance in scenarios where multiple institutions, such as banks, collaborate to generate personalized models by aggregating knowledge from different domains of the same person. FTL also leverages multiple data sources to train a model on the server. The goal is to identify commonalities across features, reducing the overall error. FTL supports both homogeneous (different samples) and heterogeneous (different features) training approaches, making it particularly valuable in healthcare applications \cite{a11}.

Each FL architecture offers distinct advantages. The horizontal architecture ensures data independence, allowing devices to update model parameters without extensive communication. Vertical architecture enables collaboration between different servers, facilitating the aggregation of diverse features. FTL leverages different data sources to identify common patterns, ultimately improving model accuracy \cite{a10, a12}.

These architectural approaches find applications across various domains where privacy and data security are paramount. For example, in the healthcare domain, by leveraging FL, healthcare institutions can collaboratively train models while preserving patient privacy, facilitating personalized healthcare services, and advancing medical research.

\subsection{Challenges}

\begin{figure}[!t]
\centerline{\includegraphics[width=0.34\columnwidth]{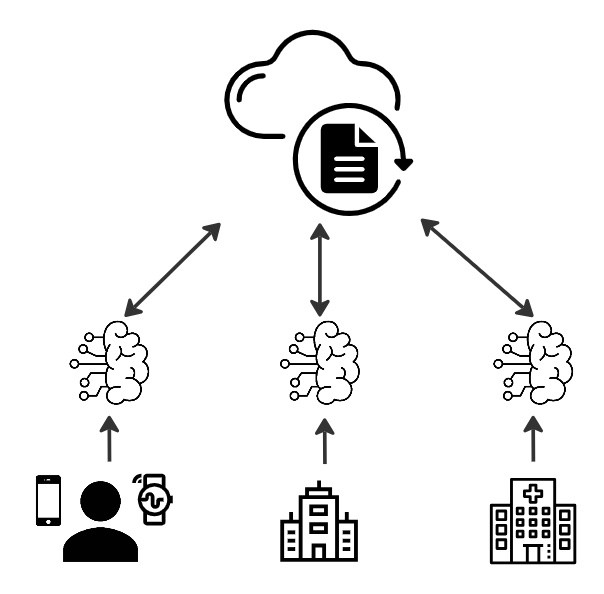}}
\caption{Federated learning with heterogeneous devices, data collection types and models  \label{fig:FLSelfMade}}
%\label{fig:TypicalFL}
\end{figure}

Federated Learning encompasses various research areas \cite{a13, a14}, including optimization \cite{a15}, communication efficiency \cite{a16}, personalization \cite{a17}, fault tolerance \cite{a18}, privacy preservation \cite{a19}, and computational efficiency \cite{a20}. Likewise the case in many methodologies, FL also has pros and downsides. These challenges mainly arise due to device, data, and model heterogeneity (Figure \ref{fig:FLSelfMade}).

\textbf{Device heterogeneity:} 
The use of different devices in data collection, such as urban sensors, smartphones, wearable devices, and autonomous vehicle sensors \cite{a21, a22}, introduces variations in sensing capabilities, memory, data processing, battery life, and communication capacity. Consequently, challenges arise, including fault tolerance, communication efficiency, and computational efficiency, due to the diverse nature of IoT devices.

\textbf{Data heterogeneity:} 
Device heterogeneity leads to non-IID (non-identically and independently distributed) data, manifesting in different labels, feature distributions, and concept drift \cite{a23}. For example, when data is collected from multiple devices, each device may have only a subset of the activity classes in Human Activity Recognition (HAR) research. Moreover, different individuals may assign different labels to the same activity. Furthermore, variations in feature distributions can occur due to individuals' unique characteristics and behaviours. Addressing these challenges requires distributed optimization techniques and personalized approaches.

\textbf{Model heterogeneity:} 
While the standard model used in basic FL architecture facilitates aggregation, it may only be suitable for some scenarios. Model heterogeneity arises from the diverse needs of client devices. For example, when constructing a model for predicting user activities using data from smartphones and smartwatches, adapting the model based on each device's capacity may be necessary. Additionally, clients may have privacy concerns that discourage sharing certain model parts. This challenge underscores the need for further research on privacy preservation and client participation.

\subsection{Federation Algorithms}

FL algorithms vary based on the model aggregation methods used in the literature. The initial algorithm proposed by Google, known as FedAvg \cite{a24}, employs the \textit{average function} for aggregation. Subsequently, several variations of averaging methods emerged, including weighted averaging \cite{a25}, one model selection, best model averaging \cite{a12}, stochastic controlled average \cite{a26}, and periodic averaging combined with quantization \cite{a27}. However, FedAvg has limitations when working with non-IID data and fails to capture fine-grained information specific to each device \cite{a28}.

Other widely used algorithms include FedMA \cite{a29}, which constructs a global model layer-wise using matching and averaging, specifically designed for neural network architectures to address data heterogeneity. FedPer \cite{a30} tackles data heterogeneity caused by non-IID distributions by providing personalized models composed of base and personalized layers. The base layers are aggregated on the server, while clients focus on representation learning in the personalized layer using transfer learning.

Moreover, the server must receive models from all participating devices to apply any aggregation method which causes latency problems. Slower clients are called stragglers, and the problem is called the straggler problem \cite{a124}. In other terms, we can define this problem as a synchronous and asynchronous learning problem. The literature proposes asynchronous \cite{a31} and semi-synchronous \cite{a32} methods to overcome the straggler problem. In addition to latency, when the server faces unbalanced and non-IID data, even these proposed methods converge slowly \cite{a33}.  

In \cite{a34}, the authors propose a partial aggregation strategy (FedPA) where the server updates its model using models from only a subset of clients determined by a reinforcement learning model. This parameter's adaptive and dynamic calculation considers data distribution and device heterogeneity, preventing accuracy loss caused by selecting an inappropriate aggregation number.

Similarly, in \cite{a35}, the authors propose the n-soft sync aggregation algorithm to address cloud idle time. This algorithm combines the benefits of synchronous and asynchronous aggregation by uploading only the models of n clients in each round, striking a balance between transmission overhead and training time. This approach ensures faster completion of training by corresponding clients.

%FedSecAgg
In \cite{a108}, authors introduce FL with Secure Aggregation (FedSecAgg), an multi-party computation (MPC)-based aggregation protocol that ensures privacy and security in FL scenarios. It addresses the challenge of privacy leakage during the aggregation process by employing secure MPC techniques. This approach allows the aggregation to be performed while preserving the privacy of individual client updates.

%FedMeta-MAML
In \cite{a109}, Federated Meta-Learning with Model Agnostic Meta-Learning (FedMeta-MAML) is proposed. FedMeta-MAML combines the principles of FL with model-agnostic meta-learning (MAML). Meta-learning enables rapid adaptation to new tasks and clients by learning a good initialization for client models. FedMeta-MAML aggregates client updates using a meta-update rule, allowing clients to learn how to learn from their local data. 

%FedMoSGD
FL with Momentum SGD (FedMoSGD) extends federated learning by incorporating the momentum technique \cite{a110}, which is commonly used in stochastic gradient descent (SGD) optimization. The momentum technique accelerates convergence and improves the robustness of the optimization process by adding a fraction of the previous update direction to the current update direction. 

Another proposed method in literature is FedProx which adresses the challenges of heterogeneity \cite{a125}. It is a generalization and re-parametrization of the popular FedAvg method, which allows for local updating and low participation to tackle communication costs. The proposed framework introduces modifications that address both system and statistical heterogeneity, providing convergence guarantees in scenarios with non-identically distributed data across devices. Empirically, FedProx demonstrates more robust convergence and improved accuracy compared to FedAvg on realistic federated datasets, especially in highly heterogeneous settings.%, where it improves absolute test accuracy by an average of 22\%.

The field of FL has witnessed advancements in aggregation methods to overcome challenges posed by non-IID data, latency, and privacy. These state-of-the-art algorithms provide valuable contributions toward enhancing FL processes' efficiency, privacy, and convergence.

\subsection{Metrics}
%Various metrics are used to evaluate and compare FL algorithms, capturing model performance, communication cost, computational power, and fairness. These metrics include accuracy and loss for model performance, global and local rounds for communication cost, convergence for computational power, and fairness for similar performance over clients \cite{a36}.

In order to evaluate and compare Federated Learning (FL) algorithms, various metrics are employed in the literature. These metrics provide insights into different aspects of FL performance and effectiveness. When assessing FL algorithms, it is crucial to consider metrics that capture model performance, communication cost, computational power, and fairness. In literature \cite{a36}, accuracy and loss for a model performance measure, number of global rounds and the amount of transmitted data for communication cost, number of local rounds and convergence for computational power measure, fairness (similar performance over clients) are used as metrics. 

Model performance in FL is evaluated using \textit{accuracy} and \textit{loss} metrics, which measure the accuracy of predictions and the deviation between predicted and actual values. Higher accuracy and lower loss values indicate better model performance in terms of prediction quality and generalization. These metrics are also used to compare FL performance with traditional ML algorithms.

\textit{Communication cost} is crucial in FL systems, with metrics like global rounds and transmitted data assessing efficiency. Minimizing the number of rounds and reducing the transmitted data volume can optimize the overall communication cost in FL systems.

\textit{Computational power} measures evaluate algorithms' efficiency in reaching stable solutions. The number of local rounds reflects the computational burden on individual client devices during training. Convergence, another important metric, evaluates the algorithm's efficiency in reaching a stable and optimal solution. Faster convergence indicates a more computationally efficient approach.

\textit{Fairness} is an essential consideration in FL, aiming to ensure equitable performance and benefits for all participating clients. Fairness metrics assess the consistency of model performance across different clients, avoiding biases or imbalances during the learning process.

%while convergence evaluates convergence and \textit{fairness}, ensuring equitable performance for all clients.

\section{State of the Art Federated Learning Software Platforms and Testbeds}
\label{sec:hardSoft}

In this section, we introduce some prominent software frameworks for building FL applications. In addition, we provide a list of popular testbeds crucial for the model deployment and evaluation on real devices.

\subsection{Software Platforms}
Frameworks provide developers and researchers with powerful tools and resources to implement FL algorithms and conduct experiments in a distributed and collaborative manner \cite{a37}. We review the commonly used FL software platforms in the literature: 

\begin{itemize}

   \item \textbf{TensorFlow (\cite{a38}):} TensorFlow Federated (TFF) is developed by Google specifically for applying deep learning over distributed data, with a particular focus on mobile keyboard prediction and on-device learning. TFF consists of two API layers: Core and FL API. The Core API offers default implementations of FL algorithms such as FedAvg and FedSgd, while the FL API allows for the definition of custom federated algorithms. TFF currently supports simulation mode which is a controlled environment rather than being executed on real distributed devices or edge nodes. It allows researchers and developers to fine-tune the algorithms and observe their behavior under various conditions without the complexity and overhead of dealing with real distributed systems.
    %https://www.tensorflow.org/federated
    %stands for "Learning with Few Labels" and 
    \item \textbf{LEAF (\cite{a39}):} LEAF is an open-source modular benchmarking framework. It comprises three components: datasets, evaluation frameworks, and reference implementations. LEAF supports various learning paradigms, including federated, meta-learning, multi-task, and on-device learning. It provides a benchmarking environment for evaluating FL algorithms.
    \item \textbf{PySyft (\cite{a40}):} PySyft is a framework focusing on secure and private deep learning applications compatible with popular deep learning libraries such as PyTorch and TensorFlow. It enables static and dynamic computations and incorporates privacy-preserving techniques into the FL workflow. PySyft leverages PyGrid, an API for managing and deploying PySyft-powered models to interact with edge devices.
    \item \textbf{Flower (\cite{a41}):} Flower is a user-friendly FL framework that provides a facility for large-scale data scenarios and is designed to support heterogeneous FL device environments. One of its significant contributions is the ability to run FL algorithms directly on edge devices. Flower also facilitates conducting experiments at the algorithmic level and considers system-related factors such as computing capability and bandwidth. It has been successfully tested in scenarios involving millions of clients.
    \item \textbf{Fate (\cite{a42}):} Federated AI Technology Enabler (Fate) is developed by WeBank and offers extensive customization options for FL algorithms. It comprises various components, including FATEFlow, FederatedML, FATEBoard, FATE Serving, Federated Network, KubeFATE, and FATE-Client. Fate supports horizontal and vertical data partitions, allowing for flexibility in data distribution. However, it currently employs a centralized implementation, where a server coordinates all the FL processes. It can be deployed in both simulated and federated modes.%https://fate.fedai.org/
    \item \textbf{OpenFL (\cite{a43}):} OpenFL is an open-source project initiated by Intel. It provides a bash-script-based framework for FL and emphasizes secure communication between clients. OpenFL consists of two main components: the abrogator and the collaborator. The framework allows for the customization of various aspects, including logging mechanisms, data split methods, and aggregation logic. However, it requires manual client-side handling.%https://github.com/intel/openfl
%https://www.apheris.com/blog-top7-open-source-frameworks-for-federated-learning

\end{itemize}

Among the listed FL software platforms, Flower and PySyft are particularly relevant for working with sensor data and edge devices, including IoT devices. Flower is a user-friendly framework supporting large-scale data scenarios and heterogeneous FL device environments. It enables running FL algorithms directly on edge devices, making it suitable for IoT devices. Flower considers system-related factors such as computing capability and bandwidth, making it a valuable choice for edge devices in sensor networks. PySyft, on the other hand, focuses on secure and private deep-learning applications. It is compatible with popular deep-learning libraries and allows for interactions with edge devices. With its privacy-preserving techniques, PySyft can be applied to IoT devices, ensuring data security during the FL workflow. These frameworks provide the necessary support and customization options to address the challenges associated with edge computing and facilitate efficient FL on resource-constrained devices.

\subsection{Testbeds}
Testbeds play a crucial role in deploying and evaluating FL on the target devices. They provide open, accessible, and reliable facilities to validate technological solutions and foster research collaboration.

One example testbed is the Fed4FIRE project\footnote{https://www.fed4fire.eu/news/innovation-through-the-fed4fire-federated-testebds/}, which aims to develop a common federation framework integrating EU and US testbeds. By enabling experiments across different domains, the project facilitates the assessment of Software Defined Networks (SDN), explores edge computing use cases, and highlights the benefits of interconnecting these testbeds.

In the realm of FL algorithms, the PTB-FLA Python Testbed  \cite{a111} offers a Python-based framework specifically designed for smart IoT devices in edge systems. It supports centralized and decentralized algorithms, providing advantages such as a small application footprint and simple installation with no external dependencies. While the current version has limitations, such as a maximum number of nodes and edges and a lack of networking support, PTB-FLA shows promise for developing FL algorithms targeting IoT devices, and future implementations aim to expand its capabilities. This testbed includes swarms of simply IoT devices (no computers) on its roadmap that may utilize MicroPython as an OS, which is becoming more prevalent in embedded systems. 

The FeatureCloud testbed\footnote{https://featurecloud.ai/} provides a controlled environment for developers to test their FL applications. It offers workflow and testbed execution modes, allowing collaborative federated training and standalone app testing. With features like simultaneous test runs, monitoring panels, configuration settings, and result storage, developers can quickly validate and manage their FL apps before deployment in a federated environment.

For realistic evaluation of FL systems, the FLAME testbed \cite{a112}, introduces a novel data partitioning scheme that distributes datasets among numerous client devices while preserving the multi-device nature of FL. This testbed generates many users, each owning a small number of labelled data samples, enabling developers to assess FL algorithms and models in a scalable setting. The FLAME testbed contributes to reproducibility and further research by open-sourcing its source code and federated HAR datasets.

These testbeds provide valuable platforms for researchers and developers to experiment, validate, and refine their technologies in edge devices and FL domains. Among these testbeds, Fed4FIRE, PTB-FLA, FeatureCloud, and FLAME, demonstrate support for edge devices in federated learning. Fed4FIRE aims to integrate various testbeds, including those supporting IoT, to enable experiments across different domains and emphasize the importance of providing open and interoperable testing facilities for IoT researchers and innovators. PTB-FLA specifically targets smart IoT devices in edge systems, offering a Python-based framework with a small application footprint suitable for smart IoTs. FeatureCloud provides a controlled environment for testing FL applications and supports IoT devices, allowing developers to validate and manage their apps before deployment in a federated environment. Similarly, FLAME supports IoT devices and creates a realistic evaluation environment for FL systems by generating users with labelledW data samples, enabling developers to assess FL algorithms in a more realistic IoT setting. These testbeds collectively contribute to advancing research and innovation in federated learning on IoT devices by offering experimentation, validation, and refinement platforms.

\section{Sensors and Edge-Sensing Devices}
\label{sec:sensorDevices}
In this study, we focus on edge sensing devices such as IoT sensors, mobile sensing devices (smartphones), wearables, and many other sensing devices that are used in the application domains of sensor-based edge devices: medical, environmental, automotive, industrial, aerospace, and agricultural applications as seen in Figure \ref{fig:Abstract}. The devices possess three fundamental abilities: context, computing, and connectivity \cite{a47}. Context refers to measuring information using sensors while extracting knowledge from the sensed data. Connectivity enables communication with other devices or central servers to facilitate knowledge transfer \cite{a47}.

\begin{figure}[!b]
\centerline{\includegraphics[width=0.66\columnwidth]{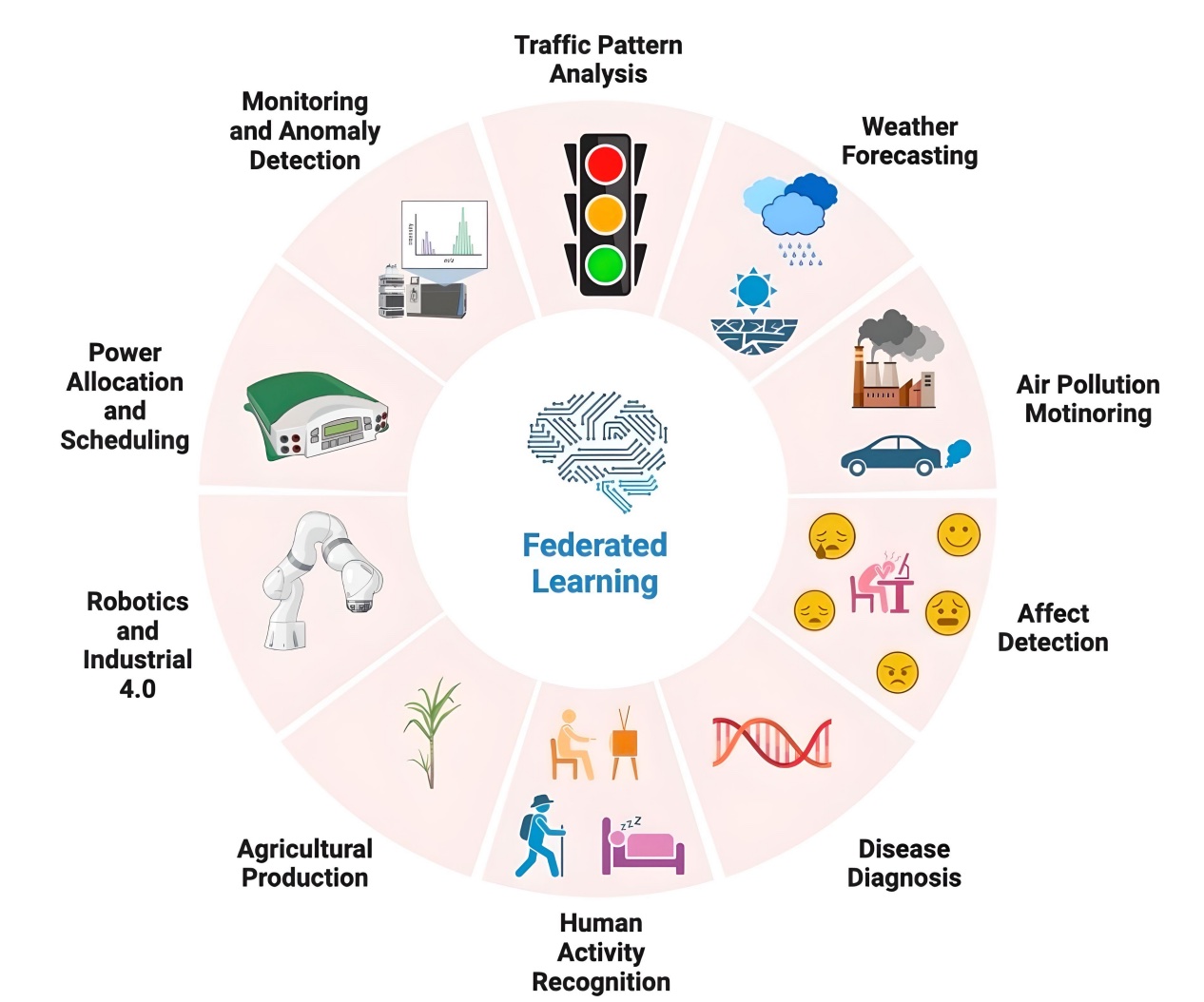}}
\caption{Domains of sensing applications covered in this study\label{fig:Abstract}}
\label{fig:Abstract}
\end{figure}
%\textcolor{red}{Bu resim bir yerden alınmamıştı değil mi, sen çizmiştin?}: yes, self-made

The emergence of sensor technology has revolutionized data collection and computation capabilities, giving rise to the concept of the Internet of Things (IoT) \cite{a44}. These small-sized sensors find applications in various domains, including health monitoring (such as heart rate fluctuations and step counts), communication and environment sensing in autonomous vehicles (using cameras and proximity sensors) \cite{a45}, as well as marketplaces, agriculture, smart manufacturing, smart buildings, and energy management \cite{a46}.

Besides dedicated IoT devices, our smartphones and wearables, such as smartwatches, also have sensors. Wearable data analysis has gained significant popularity over the past decade, with Human Activity Recognition being a typical application area \cite{a57, a58, a59}. Researchers have started focusing on more complex activities and behaviours, such as smoking, emotion, and stress recognition, rather than solely concentrating on basic activity types like walking and running.

Various sensors can be utilized in the target devices, including motion, electrodermal activity, blood volume pulse, heart rate variability (HRV), accelerometer, gyroscope, magnetometer, pressure, proximity, temperature, GPS, sound, image, water quality, chemical, gas, smoke, infrared, humidity, and optical sensors \cite{a48}.

Sensing devices' hardware properties significantly impact their performance and capabilities. The central processing unit (CPU) is a critical component, that executes instructions and calculations, and its performance is determined by clock speed, core number, and cache memory. Comparing CPU specifications across devices like smartwatches, Raspberry Pi, etc., reveals differences in processing power and computational capabilities. For instance, smartwatches have low-power CPUs suitable for basic tasks like notifications and fitness tracking while Raspberry Pi devices, with quad-core ARM-based CPUs, offer better capabilities than smartwatches and can handle lightweight desktop applications and basic gaming. However, smart phones equipped with multi-core CPUs provide significantly higher processing power, making them ideal for demanding tasks like video editing, %3D rendering, 
and machine learning, as well as running resource-intensive software efficiently. GPUs, specialized processors, enhance computational power and enable advanced graphical tasks in image processing, augmented reality, and machine learning. While some smartwatches and Raspberry Pi models may have integrated graphics capabilities as part of their SoC (System on a Chip), they are not typically equipped with dedicated GPUs like those found in PCs. Dedicated GPUs in PCs are specifically designed for high-performance graphical tasks, such as gaming, video rendering, and machine learning, providing much higher computational power for these tasks compared to integrated graphics found in smartwatches and Raspberry Pi devices. %Out of the three devices mentioned, only PCs commonly have dedicated GPUs (Graphics Processing Units)

%Memory is another crucial hardware component in devices. Devices with larger RAM capacities can handle extensive datasets and run multiple applications without performance degradation. Sensing devices like smartwatches and Raspberry Pi have lower memory capacities, requiring smaller internal storage or external MicroSD cards. These devices face challenges in resource-intensive tasks and processing large datasets, requiring developers to optimize algorithms and applications. 

Memory is another crucial hardware component in all kinds of devices. Random access memory (RAM) provides temporary storage for data that the device uses during operation. Devices with larger RAM capacities can handle more extensive datasets and run multiple applications simultaneously without experiencing significant performance degradation. Sensing devices like smartwatches and Raspberry Pi have significantly lower memory capacity than a typical PC. Smart-watches and Raspberry Pi often have 512 MB to 1 GB RAM, while PCs typically have 8 GB to 16 GB or more. Smart-watches and Raspberry Pi rely on smaller internal storage (4 GB to 8 GB) or external MicroSD cards (16 GB to 128 GB) for data storage. In contrast, PCs have more extensive internal storage options (256 GB to 1 TB or more). The limited memory specifications of sensing devices can pose challenges when performing resource-intensive tasks or processing large datasets. Developers must carefully optimize algorithms and applications to work efficiently within the memory constraints of these devices. On the other hand, PCs have ample memory and storage capabilities, making them better suited for more computationally demanding tasks and data-intensive applications.

%The target devices consist of various hardware components, including CPU, GPU, RAM, sensors, wireless connectivity, storage capacity, and power management. These properties can vary significantly across devices, models, and manufacturers. Newer generations offer more powerful CPUs, GPUs, memory capacities, and features. IoT devices have single or dual-core processors, while wearables have low-power CPUs, moderate memory, and specialized sensors. Smartphones have multi-core processors and expandable storage options. 

It is important to note that the hardware properties can vary significantly across different devices, models, and manufacturers. The specifications and capabilities of the devices are constantly evolving, with newer generations offering more powerful CPUs, GPUs, increased memory capacities, and additional features along with various devices, including IoT devices, wearables, smartphones, robots, drones, and many more. For instance, IoT devices typically possess single or dual-core processors with limited memory and storage. At the same time, wearables like smart watches are equipped with low-power CPUs, moderate memory, and specialized health-related sensors. On the other hand, smartphones feature multi-core processors, substantial memory, high-resolution cameras, and expandable storage options. Robots may vary in CPU configurations and memory capacity, relying on various sensors for perception and navigation tasks. Drones often have specialized flight controllers, moderate memory, and microSD card storage. Considering the hardware properties of these devices is essential when developing applications and algorithms to ensure optimal performance and compatibility with the target hardware platform.

In this review, our main emphasis lies on sensors commonly found on edge devices, such as motion sensors (accelerometer, gyroscope, and magnetic field sensors), location sensors (GPS and wireless interfaces), pressure sensors, thermometers, electrodermographs (EDA), electromyography (EMG), electroencephalographs (EEG), electrocardiographs (ECG), oximeters, and proximity sensors (e.g., Bluetooth). These sensors are frequently employed in various sensing applications, often in combination, resulting in the generation of multi-modal sensor data. In the medical field, sensors such as electrocardiogram (ECG) sensors are used to measure and record the heart's electrical activity, aiding in diagnosing heart conditions. Pulse oximeters, another medical sensor, monitor oxygen saturation levels in the blood. In the environmental domain, water quality sensors are utilized to assess parameters like pH levels and dissolved oxygen in bodies of water. Gas sensors are employed for indoor air quality monitoring and detecting the presence of harmful gases. Automotive applications involve light sensors that adjust headlight and dashboard display brightness based on ambient light levels and rain sensors that automatically activate windshield wipers in response to rain. In industrial settings, flow sensors measure the flow rate of liquids or gases in pipelines, facilitating flow control and monitoring. Vibration sensors are utilized to detect vibrations in machinery, identify potential faults, and prevent breakdowns. Finally, in the aerospace sector, gyroscopes are used to measure angular motion, ensuring stability and navigation accuracy. These examples demonstrate the diverse range of sensors utilized in various domains to detect and measure environmental changes, enabling precise monitoring, control, and decision-making processes. %We intentionally exclude the camera and microphone sensors frequently found on target devices in this survey and instead concentrate on other sensing devices and the kinds of data they produce.

\section{Federated Learning Applications on Sensing Devices}
\label{sec:flApp}
Small sensing devices can capture location-based or person-based data related to specific environments or behaviours. Data collected from multiple devices can be combined according to the application's requirements to gain a comprehensive understanding of situations or behaviours. This section explores the diverse application areas where FL techniques have been utilized (see Figure~\ref{fig:Abstract}). These areas include Human Activity Recognition (HAR), health monitoring, affect detection, abnormality detection, and other relevant areas where sensors on edge devices are utilized. We are inspired by the study \cite{a49} and extended it by adding different application areas. Due to the page limitations, we did not include all studies covering these application areas; instead, we provided examples of studies from these areas in the literature. In Table \ref{tab:applications}, we present an overview of the covered studies, which are explained in the following sections in detail.

\subsection{Medical applications}
In today's world, computational-based techniques are widely used in the healthcare domain, especially with image recognition, e.g., the detection of diseases, the decision of cell types according to sizes and forms.  The benefits of using these techniques are fast and accurate diagnosis and treatment \cite{a60, a113}. However, data protection regulations make it impossible to collect and analyze these data easily. In this regard, FL techniques are well suited to healthcare-related applications from the privacy perspective. In addition, intrinsically, FL is designed for a vast number of client participants.

\begin{itemize}
%\textcolor{red}{Instead of the paper, the study}
%addresses  a context-aware decision support system for cardiac health monitoring . \textcolor{red}{This sentence almost says the same thing with the prev. one, you can combine.}
    \item \textbf{Disease Diagnosis:}  The study \cite{a114} proposes an intelligent and privacy-oriented context-aware decision support system for cardiac health monitoring using physiological and activity data of the patient during rehabilitation, utilizing a federated machine learning approach for activity recognition while maintaining users' privacy. The system allows healthcare professionals to assess the health status of patients by considering multiple relevant parameters, such as heart rate, electrocardiogram (ECG) signals, and activity data. It uses the data collected from Holter monitors and smartphones equipped with accelerometers. The federated machine learning approach enables the development of a collaborative model without sharing sensitive patient data, ensuring data privacy while achieving model generalization. %This approach opens new possibilities for accurate and secure cardiac health monitoring, enhancing patient care and diagnostics.%\textcolor{red}{You may say sth related to the FL usage here.}
    \item \textbf{Affect Detection:}
    %\textcolor{red}{What are the sensors? Edge-sensing devices here? This is a good sensing app. example } 
    % \st{to improve health systems through wearable IoT devices and advanced machine learning algorithms}
    %\st{The aim was the automatic detection of stress via heart rate data.}
Affective computing \cite{a103} is an active research area for automatically monitoring a person’s mental and emotional state via physiological and physical signals. In \cite{a104}, the authors applied FL on stress data with the aim of privacy preservation. The authors applied FL on heart activity data collected with smart bands, and the targeted devices include wrist-worn wearable devices (e.g., Samsung Gear S, Empatica E4) and smartphones, with various physiological sensors such as accelerometers, electrodermal activity, heart rate, etc. The proposed federated deep learning algorithm achieves encouraging results in stress-level detection while ensuring privacy protection for health-related research conducted with widespread mobile unobtrusive devices.
Two use-case scenarios are considered: creating a general model from individual models while preserving user privacy and aggregating separately collected event data to develop an improved shared model. They applied deep learning algorithm in a federated manner and obtained better results compared to traditional learning methods.
    %cite
%1- (Feng, Rong, Sun, Guo, & Li, 2020) 
%2-   (Sozinov, Vlassov, & Girdzijauskas, 2018-->Human Activity Recognition Using Federated Learning
%Sanara
%\st{is an active research area, especially in the well-being, medical, and security domains (e.g., biometric systems) via wearable devices.}
%\textcolor{red}{Bu cümleye gerek var mı, zaten FL'in özü bu: Instead of uploading private data, the only thing uploaded to the centralized server is updated model parameters, making it secure and private.}
%\textcolor{red}{Bundan sonraki cümleler bu çalışma için mi değilse delete. Ya da bu bulletın başına almalı.}
  \item \textbf{Human Activity Recognition:} HAR aims to detect users' activities, such as sitting, standing, and walking, mostly via motion sensors. The sensors used in HAR tasks typically include triaxial accelerometers and gyroscopes found in smartphones and smartwatches. These sensors measure acceleration forces and rotational movements, allowing the models to recognize various activities such as walking, running, biking, and more. In \cite{a57}, FL models were applied for trajectory prediction. The framework utilizes various sensing devices and sensors present in mobile devices to gather location-based data. The paper introduces a group optimization method for training on local devices to achieve a better trade-off between performance and privacy. In \cite{a58}, authors applied FL to detect basic human activities, where the goal is to recognize different human activities using sensor data from smartphones or smartwatches. They obtained good accuracies but slightly lower than a centralized one. In \cite{a59}, they focused on the HAR domain and applied several aggregation algorithms (FedAvg, FedPer, and FedMA) and compared their results. %\textcolor{red}{Did they also focus on other domains? mention then.}.  %\textcolor{red}{Bu çalışma sonuçta ne elde etmiş?} %\textcolor{red}{But is this HAR?}

\end{itemize}

\subsection{Environmental applications}
FL has emerged as a powerful approach to tackle environmental challenges like climate change, air pollution, and extreme weather events.
\begin{itemize}
  \item \textbf{Air Pollution Monitoring:} FL is employed to predict air pollution using sensor networks in \cite{a115}. The utilized sensing devices are environmental monitoring stations with various sensors, which are not specified in the paper. Nevertheless, a combination of sensors is typically deployed to collect data on various environmental parameters such as particulate matter (PM), nitrogen dioxide (NO2), ozone (O3), carbon monoxide (CO), sulfur dioxide (SO2), meteorological data (temperature, humidity, wind speed, etc.), and more. The paper focuses on predicting air pollution levels, particularly Oxidant warning levels, using Convolutional Recurrent Neural Networks (CRNN). The CRNN models are trained locally on data from different spatial areas, such as cities and prefectures, and then the common parts of these models are aggregated at the server to create a global model. This federated learning framework allows for cooperative training among participants from different regions without exchanging raw data, ensuring data privacy and reducing data transmission latency.

    %The goal is to develop a collaborative model that can accurately predict air quality levels based on sensor data collected from different locations. By training the model collaboratively without sharing raw sensor data, privacy concerns are addressed while leveraging the collective knowledge of distributed sensors to improve air quality prediction. %ref: Spatially-distributed federated learning of convolutional recurrent neural networks for air pollution prediction
     \item \textbf{Weather Forecasting:} FL is utilized for weather prediction using distributed weather stations \cite{a116}. The stated paper introduces a novel machine learning approach, MetePFL, for weather forecasting on a global scale. The proposed approach utilizes a foundation model (FM) pre-trained on extensive weather data and fine-tuned by regions to capture local weather patterns. MetePFL employs prompt federated learning to train a Transformer-based FM collaboratively across participants with heterogeneous meteorological data. A spatiotemporal prompt learning mechanism is introduced to handle multivariate time series data efficiently, improving forecasting accuracy. The experiments conducted on accurate weather forecasting datasets demonstrate the effectiveness and superiority of the MetePFL approach over traditional fine-tuning and other federated learning methods. By utilizing only a fraction of the model's parameters, MetePFL achieves excellent performance while reducing inter-device communication overhead, making it a promising solution for comprehensive weather forecasting on large-scale meteorological data. %\textcolor{red}{You explained this study very well, you can improve the writing in other app areas, giving specific details about the arch's, FL and the results. Or if you are doing this in a table, refer to that and ask the reader to look at there.}

  %ref: Prompt Federated Learning for Weather Forecasting: Toward Foundation Models on Meteorological Data

    %cite
%163
    %\item \textbf{Visual security:} The possible threats can be understood by monitoring the environment. Cameras are the most used sensor for that purpose \textcolor{red}{but you mentioned that you excluded cameras in this survey?}. Any suspicious activity can be detected based on predefined feature sets. However, collecting all registered records in one place is not possible due to privacy concerns and computation costs. 

\end{itemize}

\subsection{Smart city applications}

FL techniques are also applied to integrated city solutions related to food, water, and energy \cite{a77}. Applications use data collected from cameras, temperature sensors, and traffic sensors to understand human activities, population density, and pollution \cite{a78}. 

\begin{itemize}

    \item \textbf{Traffic Pattern Analysis:} FL is employed for traffic flow prediction in smart cities \cite{a117}. The paper \cite{a117} presents FedGRU, a novel privacy-preserving traffic flow prediction algorithm based on FL. The goal is to develop a collaborative model that can accurately predict traffic patterns and congestion levels using data from various sources, such as traffic cameras and sensors. The proposed algorithm uses a secure parameter aggregation mechanism. The paper introduces a joint-announcement protocol to handle large-scale scenarios, randomly selecting a subset of organizations for each training round, effectively mitigating communication challenges. Empirical case studies demonstrate that FedGRU achieves comparable prediction accuracy to centralized models while maintaining robust privacy protection. An ensemble clustering-based approach is proposed to enhance model performance by leveraging spatiotemporal correlations among organizations' data.

 %ref: Privacy-preserving traffic flow prediction: A federated learning approach
    \item \textbf{Smart Transportation:} 
Transportation is composed of two parts: vehicular traffic planning and resource management \cite{a79}. Traffic planning is essential for traffic prediction and congestion minimization. In \cite{a80}, authors replaced centralized ML models with FL models for traffic prediction cases and ran the model on edge devices. As data, they utilized traffic flow, weather, and road geometry. In \cite{a82}, a traffic simulator is designed with FTL to guide Reinforcement learning agents' (FTRL) driving. The proposed framework provides collision avoidance as well.

\end{itemize}

\subsection{Industrial applications}
Processing industrial data with AI techniques makes possible high-precision location detection, the creation of early warning systems, and the detection of abnormal behaviors. Thus, with smart security, pre-warnings, and post-event analysis can be performed. The manufacturing process comprises modeling, monitoring, prediction, and control stages \cite{a86}. The smart industry is a concept about integrating ML techniques into this process.
%Adaptation of existing algorithms for a large amount of collected data in a privacy-preserved way is made via FL for the problem of multi-site fMRI classification with a privacy-preserving strategy in \cite{a71}. \textcolor{red}{(Is this an industrial app, fmri?)}

\begin{itemize}
    \item \textbf{Predictive Maintenance:}
FL is applied to develop models for predictive maintenance in industrial settings in \cite{a118}. The paper evaluates FL aggregation methods for predictive maintenance and quality inspection in Industry 4.0, ensuring data privacy. Four strategies, FedAvg, FedProx, qFedAvg, and FedYogi, are explored. Data distributions significantly impact FL performance. The authors introduce a real-world quality inspection dataset (FLADI) with diverse product variants as clients. FL relies on decentralized training, where clients train local models and share updates with a central server. Targeted sensing devices include machines, production units, sensors, and intelligent factories. FL's suitability varies based on data distribution and feature heterogeneity among clients.

\item \textbf{Monitoring and Anomaly Detection:}
 In \cite{a87,a88}, authors use FL for monitoring the environment collaboratively using various sensors to detect defects. In \cite{a89}, authors developed an anomaly detection system using FL for IoT devices equipped with an ammeter, water meter, and camera sensors. In \cite{a90}, it is used for attack detection for Unmanned Aerial Vehicles, which can be considered anomaly detection.

  \item \textbf{Robotics and Industry 4.0:} FL is used for allocating devices for real-time data processing in robotics. For instance, in \cite{a91}, authors used FL to decrease the communication time, facilitating learning on the device. In \cite{a92, a93}, FL is applied for learning an imitation scheme. Each device runs the NN model independently and shares the parameters with the server, which increases the global knowledge knowledge leading to more accurate results.
  \end{itemize}

\subsection{Aerospace applications}
The aerospace industry is equipped with cameras (for visual data), LiDAR (Light Detection and Ranging) sensors (for 3D mapping and obstacle detection), GPS (Global Positioning System) receivers (for location information), and inertial sensors (gyroscopes and accelerometers for attitude and motion tracking), is continually seeking innovative ways to enhance the safety, efficiency, and reliability of aircraft operations. Key to achieving these objectives is the development of advanced technologies that enable real-time monitoring of aircraft health and optimizing critical resources \cite{a121}. 

\begin{itemize}
    \item \textbf{Aircraft Health Monitoring:} FL is employed for aircraft monitoring using distributed sensor networks in \cite{a119}. The study addresses the challenge of intelligent fault diagnosis in aerospace equipment, focusing on bearings as critical components. Data-driven approaches have shown promise, but the scarcity of labelled data and data distribution discrepancies among different equipment hinder model training. The authors propose a novel method using Clustering Federated Learning (CFL) with a self-attention mechanism to overcome this. Sensors collect vibration signals from various devices, and the CFL approach clusters clients with similar data distributions. The proposed method achieved superior fault diagnosis results compared to other methods, demonstrating its effectiveness in leveraging distributed data for improved safety and maintenance of aerospace equipment.

     \item \textbf{Power Allocation and Scheduling:} FL is applied to optimize power and schedule in the aviation industry. The goal is to develop a shared model to improve fuel efficiency, reduce emissions, and enhance flight safety. Airlines or aviation companies train local models using their flight data, and the models' updates are aggregated to create a global model. By collaboratively learning from diverse flight patterns, FL enables optimized flight planning without sharing sensitive flight information. The paper \cite{a120} presents a novel framework for distributed FL in a swarm of unmanned aerial vehicles (UAVs). Each UAV trains a local FL model based on its data, and a leading UAV aggregates these models to generate a global one, optimizing convergence using joint power allocation and scheduling. The study analyzes how wireless factors impact FL performance, highlighting the effectiveness of the proposed approach.

     \end{itemize}

\subsection{Agricultural applications}
    The agricultural applications exemplify how integrating smart sensors and FL revolutionizes the state of the art. By empowering farmers with automated disease detection, crop monitoring, and efficient resource management, these technologies play a pivotal role in maximizing agricultural productivity and sustaining livelihoods in the agricultural sector.

    \begin{itemize}
    \item \textbf{Crop Disease Detection:} FL is utilized for collaborative crop disease detection in \cite{a122}. The method involves training machine learning algorithms across a decentralized network of edge devices or servers, each retaining its local data samples. By utilizing image datasets of crop diseases and employing convolutional neural networks (CNNs), FL enables the creation of a powerful deep-learning model for disease identification. This approach eliminates the need for costly and continuous professional inspection, making disease detection more accessible and cost-effective for farmers, even in remote areas. FL's privacy-preserving nature ensures farmers can contribute their crop images without sharing raw data. Implementing FL for crop disease detection promises to enhance agricultural productivity, ensuring food security and economic growth for farming communities and the entire country.

     \item \textbf{Agricultural Production:} Another application area of FL is agricultural production \cite{a123}. In this study, authors utilize various smart sensors for agricultural monitoring and control. The specific sensors used include optical sensors, mechanical soil sensors, location sensors, electrochemical sensors, dielectric soil moisture sensors, airflow sensors, and electronic sensors. These sensors are deployed in smart agricultural plots to sense crucial parameters such as temperature, pressure, humidity, soil moisture, crop health (infections and growth level), and climate conditions. The target device for this application is the smart sensor system, which integrates with IoT technologies and cloud computing for decentralized data processing and global actuation. %The smart sensors continuously monitor the environment and collect data, which is transmitted to a cloud-based platform for storage and analysis. 
     The sensor system, using FL, collaborates with the cloud server and other sensor devices to optimize sensor controls based on past accumulated data. By modifying the sensor operations according to the validated and optimal data, the smart sensors enhance agricultural production and maximize productivity in various agricultural plots.

     \end{itemize}

\begin{table}[!h]
\caption{A summary of Sensing Applications on Edge Devices with Federated Learning }% with FL Algorithm, Methodology, Sensors, and Target Edge-Sensing Devices}
\label{tab:applications}
%\begin{fntable}
\centering
\resizebox{\textwidth}{!}{
\begin{tabular}{|l|l|l|l|l|}
\hline
\textbf{Application} & \textbf{FL Algorithm} & \textbf{FL Methodology} & \textbf{Sensors} & \textbf{Target Edge-Sensing Devices} \\
\hline
Cardiac Health Monitoring (\cite{a114}) & FedAvg & Horizontal & Accelerometer & Smartphones \\
\hline
Biomedical Monitoring (\cite{a104}) & FedAvg & Horizontal, FTL & PPG & Wearable IoT devices \\
\hline
Human Activity Recognition (\cite{a58}) & FedAvg & Horizontal & Accelerometer, gyroscope & Wearables (Smartphones/watches) \\
\hline
Evaluation of FL Algorithms (\cite{a59}) & FedAvg, FedPer, FedMA & Horizontal & Accelerometer, gyroscope & Smartphones \\
\hline
Air Pollution Prediction (\cite{a115}) & Convol. Recurrent Network, & Vertical & Temperature, humidity, wind & Environmental monitoring \\
 & Averaging Aggregation & & speed, atmospheric sensors & stations \\
\hline
Weather Forecasting (\cite{a116}) & MetePFL & Horizontal & Meteorological sensors & Weather stations \\
\hline
Traffic Flow Prediction (\cite{a117}) & FedGRU & Horizontal & Traffic cameras and sensors & Edge-sensing devices for traffic flow \\
\hline
%Vehicular Networks \cite{a81} & FedGRU & Horizontal & Vehicle sensor data & Connected vehicles \\
%\hline
Autonomous Driving (\cite{a82}) & FTRL & Horizontal & LiDAR & NVIDIA Jetson TX2 \\
\hline
Predictive Maintenance (\cite{a118}) & FedAvg, FedProx, qFedAvg, & Horizontal & Air pressure, temperature, & Edge devices in industrial settings \\
 & FedYogi & & speed, force & \\
\hline
Quality Inspection (\cite{a89}) & "cli-max greedy" for FL & Horizontal & Ammeters, water meters, & Computing-enabled mobile nodes, \\
 & & & cameras & intelligent grid terminals \\
\hline
Aircraft Health Monitoring (\cite{a119}) & Clustering FL (CFL) & Horizontal, & Vibration sensors & Aerospace's machinery \\
 & & Vertical, FTL & (accelerometers) & \\
\hline
Flight Path Optimization (\cite{a120}) & FedAvg & Horizontal & GPS, IMU, cameras, LiDAR, & UAVs \\
 & & & other environmental sensors & \\
\hline
Crop Disease Detection (\cite{a123}) & FedAvg & Horizontal & Soil, location, electrochemical, & Agricultural equipment, drones, \\
 & & & airflow, electronic sensors & or IoT devices \\
\hline
\end{tabular}
}
%\end{fntable}
\end{table}
%\end{landscape}

\section{Discussion}
\label{sec:discussion}

The discussion section introduces several points that have emerged from the literature review on applying FL methodologies to sensing devices.

Firstly, it is noteworthy that most studies in the literature predominantly focus on synchronized FL, where training and model transfer occur in a synchronized manner \cite{a35}. However, the reality of systems and data heterogeneity often leads to asynchronous training and model transfer, posing challenges in scaling federated optimization. To enable efficient and scalable federated optimization, future research should explore techniques and algorithms that effectively handle the asynchrony inherent in FL settings.

Another key point to consider is the assumed labelled data in FL when using supervised algorithms, while on-device data is typically unlabelled \cite{a11}. This presents a challenge when applying FL methodologies directly to on-device sensing tasks. Future research should concentrate on developing techniques that can effectively leverage the unlabelled on-device data and explore strategies for incorporating label acquisition processes within the FL framework.

It is worth noting that many FL studies in the literature have primarily focused on simulation environments. While simulations provide controlled and reproducible settings, there is a need for more studies that adapt FL methodologies to real-world applications. Real-world deployment of FL on sensing devices introduces additional challenges, including communication constraints, privacy concerns, and device heterogeneity. Further research should aim to bridge the gap between simulation and real-world deployment, ensuring the practicality and effectiveness of FL methodologies in real-world sensing applications.

The impact of wireless communication and resource efficiency is a significant aspect when applying FL methodologies to sensing devices. While FL minimizes data transmission by uploading local updates instead of raw data, the size of the trained model parameters, especially with Deep Neural Networks (DNNs), can still be substantial. This poses a challenge for resource-constrained, battery-operated devices that may require numerous wireless communication rounds and iterations.

Efficient wireless communication is crucial in distributed machine learning to address the limitations of resource-constrained devices. Various methods can be employed to achieve communication efficiency, such as reducing the size of model updates, optimizing communication frequency, and employing selective client participation. These approaches aim to minimize the data exchanged during model updates, reducing communication overhead.

Training DNN models on resource-constrained clients, such as wearable and IoT devices, presents a significant challenge. These devices often have limited computation power, memory, and energy resources. The computational cost of running the full DNN model can be prohibitively high for such devices. Therefore, it is essential to design computationally efficient algorithms, considering the limited processing capabilities of these devices. Optimizing processing speed becomes crucial as it directly impacts the algorithm's throughput, latency, and response time.

Furthermore, personalization is a key aspect that holds potential in applying FL methodologies to sensing devices. The ability to tailor the learning process and model parameters according to individual device characteristics and user preferences can significantly enhance the performance and user experience of FL-based applications. One approach to personalization in FL is to adapt the training process to individual sensing devices' specific capabilities and constraints. As mentioned, different devices possess varying computational power, memory capacity, and energy resources. By considering these device-specific factors during the training phase, it is possible to optimize the learning process and model updates to suit the capabilities of each device. This personalized training approach can improve performance and energy efficiency by leveraging individual devices' strengths and limitations.

Another aspect of personalization in FL is user-centric customization. Sensing devices are often intimately connected to their users, capturing personal data and insights. By incorporating user preferences and characteristics into the learning process, FL can create personalized models that cater to specific user needs. This can be particularly relevant in applications such as health monitoring or affective computing, where individual variations and preferences play a significant role. Personalizing the FL models based on user-specific data can greatly enhance the accuracy and relevance of the generated insights.

Privacy is a critical concern in FL, especially when dealing with personal data. However, personalization and privacy can coexist through the adoption of privacy-preserving techniques. Techniques such as differential privacy, federated encryption, and secure aggregation can be employed to protect sensitive user data while still enabling personalized learning. By ensuring that personalization is conducted privacy-consciously, FL can strike a balance between customization and data protection, gaining user trust and fostering broader adoption.

Moreover, personalization in FL can extend beyond individual devices and users to cater to specific application domains. Different sensing applications, such as medical, environmental, automotive, industrial, aerospace, and agricultural, may have unique requirements and objectives. By tailoring the FL methodologies and model architectures to the specific demands of each domain, personalized FL solutions can be developed to address the distinct challenges and opportunities presented by different application contexts.

\section{Conclusion}
\label{sec:conclusion}

This paper has provided an overview of the current state-of-the-art FL methodologies applied to sensing devices, focusing on IoT sensors, mobile devices, and wearables. By examining the challenges and opportunities in applying FL to sensing tasks, the paper aims to inspire further research in this domain.

The discussion section has raised several important points for consideration. It highlighted the need to address the asynchrony in FL, leverage unlabelled on-device data, explore FL applications in Human Activity Recognition (HAR) and affective computing, and adapt FL methodologies from simulation environments to real-world applications on sensing devices.

Efficient wireless communication, resource efficiency, and personalization are crucial when applying FL methodologies to sensing devices. Future research should focus on developing wireless communication techniques that minimize data exchange, designing algorithms with resource constraints in mind, and enabling privacy-preserving personalization. By addressing these challenges, FL methodologies can be effectively applied in real-world sensing applications, considering sensing devices' unique constraints and requirements.

 Moreover, the discussion emphasizes the need for studies that bridge the gap between simulation and real-world deployment, ensuring the practicality and effectiveness of FL methodologies in real-world sensing applications. By adapting FL methodologies to real-world settings, researchers can address additional challenges, such as communication constraints, privacy concerns, and device heterogeneity, and validate the performance of FL in diverse sensing domains.

In conclusion, this paper aims to provide a comprehensive review for researchers and practitioners interested in applying FL methodologies to edge sensing devices to contribute to developing intelligent and efficient systems in diverse application domains by highlighting the specific challenges and opportunities associated with these devices.

\section*{Acknowledgment}
%{\color{blue}. }
Tübitak Bideb 2211-A academic reward is gratefully acknowledged. This research has been supported by the Boğaziçi University Research Fund, project number: 19301P.

%\reftitle{References}
\bibliographystyle{elsarticle-num-names}

\bibliography{main}

%=====================================
% References, variant B: internal bibliography
%=====================================

%\end{thebibliography}

\end{document}